\documentclass{article}

\usepackage{arxiv}

\usepackage[utf8]{inputenc} %
\usepackage[T1]{fontenc}    %
\usepackage{hyperref}       %
\usepackage{url}            %
\usepackage{booktabs}       %
\usepackage{nicefrac}       %
\usepackage{microtype}      %
\usepackage{graphicx}
\usepackage[numbers,sort&compress]{natbib}
\usepackage{doi}

\usepackage{newfloat}
\usepackage{listings}

\usepackage{amsmath}
\usepackage{amsfonts}
\usepackage{amssymb}
\usepackage{multirow}
\usepackage{enumitem}

\newcommand{\fire}{\raisebox{-0.2em}{\includegraphics[height=1em]{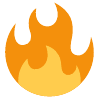}} }
\newcommand{\ice}{\raisebox{-0.2em}{\includegraphics[height=1em]{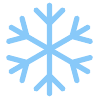}} }

\newcommand{\firenospace}{\raisebox{-0.2em}{\includegraphics[height=1em]{icons/fire.pdf}}}
\newcommand{\icenospace}{\raisebox{-0.2em}{\includegraphics[height=1em]{icons/ice.pdf}}}

\usepackage[capitalize,noabbrev]{cleveref}

\title{Eliciting Chain-of-Thought Reasoning for Time Series Analysis using Reinforcement Learning}

\usepackage{authblk}

\author[1]{Felix Parker\thanks{\texttt{fparker9@jhu.edu}}}
\author[1]{Nimeesha Chan}
\author[1]{Chi Zhang}
\author[1]{Kimia Ghobadi}
\affil[1]{Center for Systems Science and Engineering, Johns Hopkins University, Baltimore, MD 21218}

\renewcommand{\headeright}{}
\renewcommand{\undertitle}{}
\renewcommand{\shorttitle}{COUNTS}

\hypersetup{
pdftitle={Eliciting Chain-of-Thought Reasoning for Time Series Analysis using Reinforcement Learning},
pdfsubject={},
pdfauthor={Felix Parker},
pdfkeywords={Time Series, LLMs, Multimodal, Temporal Reasoning},
}

\date{}

\newcommand{\ul}[1]{\underline{#1}}

\begin{document}

\maketitle

\begin{abstract}
Many real-world time series problems require more than pattern matching: solving them often involves multi-step reasoning, domain knowledge, and unstructured context, such as interpreting ECG signals in light of clinical factors or forecasting conditioned on external events. Although reinforcement learning has recently enabled large language models (LLMs) to produce effective chain-of-thought reasoning in domains like math and code, LLMs remain weak on numerical time series due to limited time series perception and a lack of learned temporal reasoning strategies.
We introduce \textbf{C}hain-\textbf{O}f-Thought for \textbf{U}nderstanding \textbf{N}umerical \textbf{T}ime \textbf{S}eries (COUNTS), a framework that equips a pretrained LLM to understand time series inputs and to reason explicitly over diverse time series tasks using RL with verifiable rewards. COUNTS first learns a high-fidelity discrete time series representation via a residual vector-quantized variational autoencoder, yielding tokens that integrate seamlessly with the LLM vocabulary. We then train the model in two stages: supervised fine-tuning on interleaved text--time-series instructions to learn the representation, followed by a variant of Group Relative Policy Optimization on problems with known answers, where rewards are computed from task metrics (e.g., forecasting error and classification accuracy) and adherence to a format with explicit thinking and answer sections.
Across multiple benchmarks, this RL stage encourages intermediate reasoning and consistently improves performance over supervised baselines, especially on tasks requiring contextual integration and multi-step temporal inference.
\end{abstract}

\section{Introduction}

Many critical real-world problems require complex reasoning about numerical time series data, combining pattern recognition with contextual understanding and domain knowledge. Medical diagnosis from vital signs and ECG traces demands not just anomaly detection but causal reasoning about physiological states and other factors; weather forecasting requires integrating multiple sensor streams with physical models and historical patterns; and financial market analysis needs to synthesize price movements with news events and economic indicators. These tasks traditionally require domain experts often working alongside prediction models and in large teams -- a process that is time-intensive, expensive, and difficult to scale.

While traditional time series models excel at specific tasks like forecasting or classification, they struggle when problems demand higher-level reasoning or integration of unstructured contextual information. A state-of-the-art forecasting model can predict tomorrow's temperature but cannot explain why a sudden drop might indicate an approaching storm system, nor can it incorporate weather advisories or satellite imagery descriptions into its predictions. Similarly, anomaly detection models can flag irregular heartbeats but cannot reason about patient history, medications, or symptoms to determine clinical significance.

Large language models (LLMs) offer compelling advantages for such complex reasoning tasks: they can process unstructured context, possess broad domain knowledge, and -- when trained with reinforcement learning (RL) for chain-of-thought (CoT) reasoning -- demonstrate sophisticated problem-solving capabilities. Reasoning models show that RL-trained LLMs can achieve expert-level performance on mathematics and programming by learning to ``think before answering.'' However, LLMs remain remarkably poor at understanding numerical time series data. Both text representations (e.g., ``0.72, 0.85, 0.91...'') and visual encodings lose crucial numerical precision and temporal relationships, while LLMs' training data contains little time series content and no explicit reasoning about temporal patterns.

We propose \textbf{C}hain-\textbf{O}f-Thought for \textbf{U}nderstanding \textbf{N}umerical \textbf{T}ime \textbf{S}eries (COUNTS), a framework that bridges the gap between LLMs' reasoning capabilities and complex time series analysis tasks. First, we augment a pretrained LLM with improved time series perception capabilities by developing a high-fidelity tokenization method using a residual vector-quantized variational autoencoder (RVQ-VAE). This creates a discrete vocabulary for time series patches that preserves numerical precision while enabling seamless integration with the LLM's existing vocabulary. We then fine-tune the model on diverse time series tasks, teaching it to process interleaved sequences of text and time series tokens.

Building on this foundation, we apply reinforcement learning to train the model to generate explicit reasoning chains before producing answers. During RL training, COUNTS solves time series problems (forecasting, classification, anomaly detection) and receives rewards based on answer correctness, automatically computed from ground truth labels, and proper formatting of its reasoning process. This verifiable reward signal guides the model to discover reasoning strategies that improve task performance, from identifying seasonal patterns to comparing current observations against historical baselines. Unlike human preference learning, these objective rewards ensure the learned reasoning directly optimizes for task success.

Time series analysis is particularly well-suited for RL-based reasoning development because of abundant labeled datasets with clear evaluation metrics -- a verifiability property it shares with mathematics and coding domains where RL has proven transformative. To our knowledge, COUNTS is the first framework to leverage RL for training LLMs on time series reasoning tasks.

In summary, this work makes the following contributions:
\begin{enumerate}[leftmargin=*]
    \item \textbf{High-Fidelity Time Series Tokenization}: We develop a Residual Vector-Quantized VAE for time series signals that represents patches as a sequence of discrete tokens, balancing reconstruction fidelity with vocabulary size and representation quality. We train this model on a large corpus of diverse data, resulting in a universal tokenizer for numerical time series data.
    \item \textbf{Data Collection and Synthesis}: We have collected a large corpus of labeled time series data from over 15 different sources spanning diverse domains, including weather, financial, and medical datasets. We have also generated synthetic time series question-answering data using a variety of methods. This data is converted to instruction-response and question-answer pairs to facilitate training.
    \item \textbf{Unified Model for Time Series Analysis}: We train an LLM, augmented with the time series tokenizer, on interleaved multimodal sequences of text and time series tokens to enable the LLM to understand and generate numerical time series data.
    \item \textbf{RL Framework for Time Series Reasoning}: We introduce the first reinforcement learning framework specifically designed to train LLMs to perform explicit chain-of-thought reasoning on time series tasks. By using verifiable task metrics as reward signals (e.g., forecasting accuracy, classification correctness), COUNTS learns to generate reasoning strategies that directly optimize for objective task success, without requiring human feedback.
    \item \textbf{Strong Empirical Results}: We achieve state-of-the-art or competitive performance across multiple time series benchmarks, with particularly strong gains on tasks requiring reasoning and contextual understanding. We perform a thorough analysis of the model to demonstrate the effectiveness of the RL training process and time series tokenization.
\end{enumerate}
We will make the code and data publicly available to facilitate future research.

\section{Related Works}

The application of LLMs to time series analysis is a burgeoning field, driven by the potential to leverage LLMs' reasoning and contextual understanding capabilities \citep{moment, llmtime}. However, significant challenges remain, primarily concerning data representation and the development of robust reasoning mechanisms specific to temporal dynamics. Our work, COUNTS, builds upon recent advances in multimodal LLMs, discrete representation learning, chain-of-thought reasoning, and reinforcement learning for LLMs.

\paragraph{LLMs for Time Series: Representation}

Initial efforts to adapt LLMs for time series faced fundamental hurdles. Standard LLMs struggle with numerical sequences due to inefficient tokenization of continuous values into discrete text tokens, disrupting inherent mathematical properties and leading to poor performance on basic temporal tasks \citep{pitfalls_tokenization, ts_reasoning_eval}. Several paradigms have emerged to address this:
\textit{Direct Text Encoding:} Approaches like LLMTime \citep{gruver2023large} represent time series as sequences of numerical text tokens. While allowing direct use of pre-trained LLMs and text context, this suffers from inefficiency (multiple tokens per value) and poor numerical fidelity, hindering quantitative reasoning \citep{pitfalls_tokenization}.
\textit{Visual Encoding:} Methods like VL-Time \citep{liu-etal-2025-picture} convert time series into images (plots), processed by vision-language models (VLMs) with reasoning capabilities. This leverages powerful visual pattern recognition but inevitably loses numerical precision and fine-grained detail crucial for many analyses.
\textit{Specialized Encoding:} Rather than convert time series to a data modality that existing LLMs can understand, many methods have been proposed to build specialized encodings of time series signals and integrate them into LLMs. Methods like Time-LLM \citep{timellm} divide time series signals into patches, encode each, and align them with the text representations of a pretrained LLM. This enables good numerical fidelity and information density, at the cost of additional training.

\paragraph{LLMs for Time Series: Beyond Forecasting}
While there has been a lot of attention on applying LLMs to time series tasks, almost all of the work thus far has focused on time series forecasting, with limited text inputs and no text outputs. This limits the potential value of incorporating pretrained LLMs, which excel on tasks that involve natural language.
Recently more work has explored tasks that involve understanding and reasoning about time series data in natural language \citep{williams2024contextkeybenchmarkforecasting,kong2025time,liu2024time}, but this has largely focused on applying existing LLMs to new multimodal datasets, where they get mediocre results. Methods that train LLMs to understand this data are needed.
ChatTime~\citep{wang2025chattime} discretizes continuous time series values into tokens, enabling unified text–time series input and output within an LLM, and trains the LLM for forecasting and question-answering.
ChatTS~\citep{xie2024chatts} employs an MLP encoder to project time series patches into an LLM's embedding space and trains the model using a large synthetic dataset for question-answering.
While these methods have successfully trained LLMs that can understand and analyze time series data, they do not beat specialized methods for time series tasks and do not have strong high-level reasoning abilities that allow them to solve challenging problems.

\subsection{Reinforcement Learning for Reasoning}

Explicit chain-of-thought reasoning is extremely powerful for helping LLMs to correctly solve challenging problems, as evidenced by numerous works over the past year, including DeepSeek-R1~\citep{deepseekr1}.
These methods use reinforcement learning to train LLMs to generate sequences of tokens (a chain-of-thought) before giving the final answer, which has proven highly effective in verifiable domains -- where final outcomes can be reliably checked \citep{raschka_rlvr}. RL algorithms like Group Relative Policy Optimization (GRPO) \citep{shao2024deepseekmath} are used to train the LLM policy to maximize these rewards, thereby learning robust reasoning strategies.
COUNTS leverages this paradigm by recognizing that many core time series tasks (forecasting, classification, multiple-choice question-answering) are inherently verifiable, and there are many public datasets with the necessary labels.
This training enables COUNTS to reason through problems and solve them with better performance, even without any human-provided reasoning examples, allowing it to perform significantly better than existing time series LLMs.

\section{Methodology}

This section details the methodology employed in COUNTS, our framework for training LLMs to perform time series analysis tasks using explicit chain-of-thought reasoning. Our approach comprises three main components: (1) a novel time series encoding and discretization scheme using a Residual Vector-Quantized Variational Autoencoder (RVQ-VAE) to transform time series into a sequence of discrete tokens; (2) integration of these time series tokens into a pretrained decoder-only LLM; and (3) a two-phase training process involving supervised fine-tuning (SFT) followed by reinforcement learning using Group Relative Policy Optimization (GRPO) to elicit CoT reasoning for solving time series tasks.

\begin{figure*}[htb]
    \centering
    \includegraphics[width=1.0\linewidth]{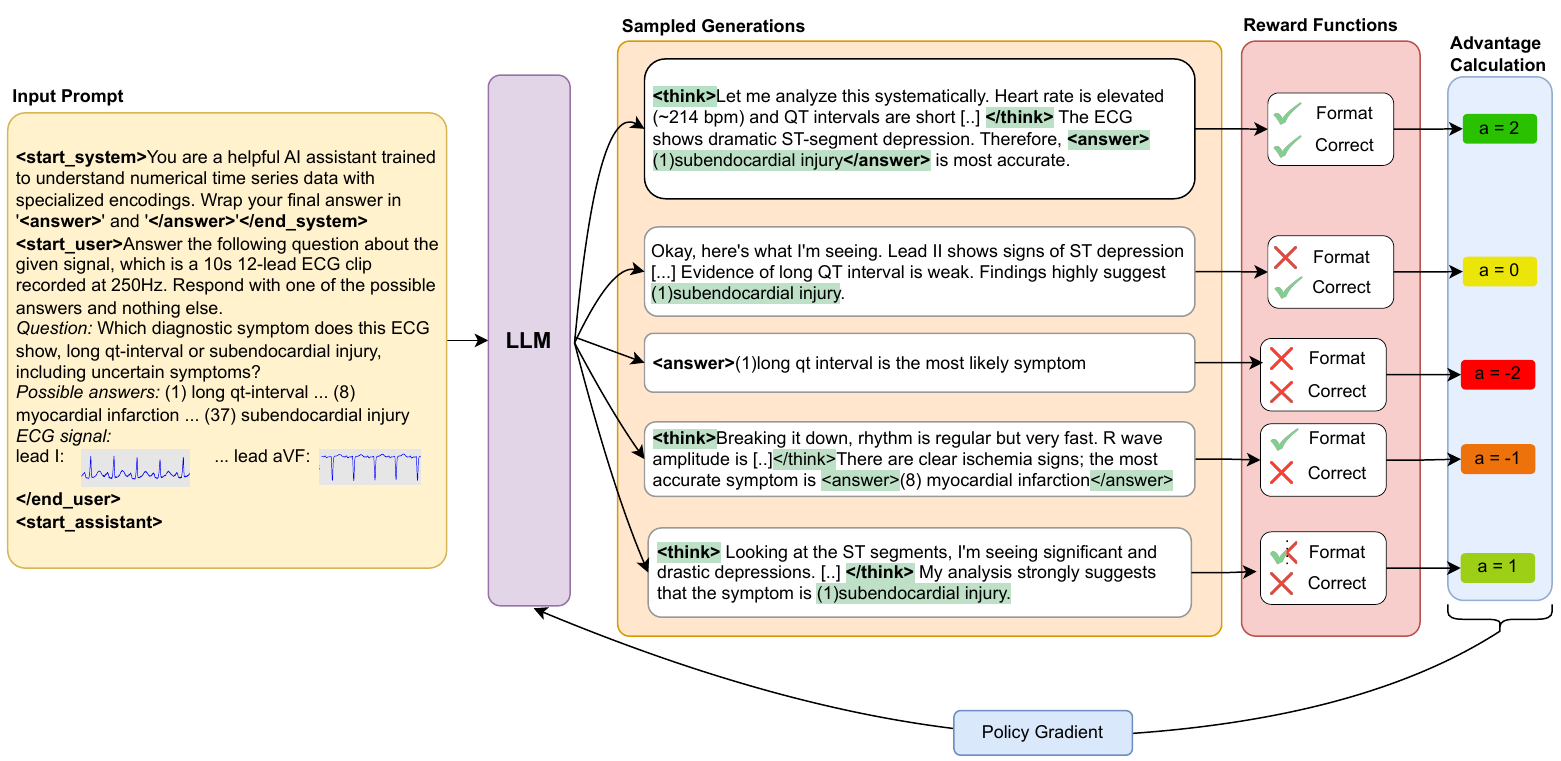}
    \caption{\small An LLM generates multiple sampled responses to an input prompt asking for ECG time series interpretation. Each response is evaluated by reward functions that assess format compliance (proper use of tags) and diagnostic correctness, with correct components highlighted in green. The resulting advantage scores are calculated on a color gradient (a = -2 to a = 2, red to green), which guide policy gradient updates.}
    \label{fig:diagram:rl}
\end{figure*}

\subsection{Time Series Tokenization}

To enable an LLM to process numerical time series data effectively we convert raw time series signals into sequences of discrete tokens that can be integrated into the LLM's vocabulary using a specialized time series tokenizer that we introduce in this section.
The tokenization process involves dividing a time series into patches, scaling each patch for numerical stability, embedding them using a simple encoder model, and finally discretizing using residual vector quantization.
The tokenizer encoder, and a corresponding decoder model, are trained using a Variational Autoencoder objective on a large collection of diverse time series data.

\paragraph{Tokenizer Encoder}
Given a univariate time series, we first divide it into non-overlapping patches of 64 time points each.
Each patch is re-scaled independently to handle the wide range of values commonly encountered in real-world time series data. Scaling the data is crucial for avoiding training instability, and disentangles shape from scale, allowing the encoder to learn better representations. Explicit scale information can also be very useful downstream for the LLM.
For each patch the scaling factor is computed by taking the absolute value of each point, identifying the maximum value, and rounding this maximum to the nearest power of 2, clipped between $2^{-10}$ and $2^{36}$. The values in the patch are then divided by this scaling factor.
While this approach is unconventional, it allows the scaling factors to be quantized into a finite, discrete vocabulary using a log transform, effectively handles an extremely wide range of scales, and only uses a single value. This means it can be easily transformed into a token representation for the LLM.

The encoder and decoder models utilize a multi-layer perceptron architecture incorporating SwiGLU layers~\citep{shazeer2020glu}, residual connections, and RMSNorm~\citep{zhang2019root}.
The encoder consists of 6 layers with hidden dimension 512, and processes the scaled patch through successive transformations to produce a continuous 128-dimensional embedding. The decoder mirrors this architecture in reverse, reconstructing the original patch from the quantized representations.

After encoding, each patch's continuous embedding must be discretized to create tokens compatible with the LLM vocabulary. We employ Residual Vector Quantization (RVQ)~\citep{zeghidour2021soundstream}, a hierarchical quantization scheme that progressively refines the representation across multiple codebooks. Unlike standard vector quantization which maps each embedding to a single discrete code, RVQ decomposes the embedding into a sum of multiple codebook vectors, enabling much higher reconstruction fidelity without requiring exponentially large codebooks.

This multi-stage quantization is essential for time series data where subtle numerical variations often carry critical information. Standard VQ-VAE approaches suffer from significant information loss when compressing a 64-point patch into a single discrete token. By using three sequential quantization stages, each refining the residual error from the previous stage, RVQ achieves near-perfect reconstruction while maintaining a tractable vocabulary size. This preservation of fine-grained numerical details is crucial for downstream tasks like anomaly detection or precise forecasting where small deviations matter.

Our specific RVQ configuration uses three quantization levels ($L=3$), with each level having a codebook of 2048 vectors. This results in each patch being represented as a sequence of three discrete tokens. Combined with the scale token that is prepended to capture the magnitude information, each 64-point time series patch is ultimately encoded as exactly 4 discrete tokens that can be seamlessly integrated into the LLM's vocabulary.

\paragraph{Tokenizer Training}

\begin{figure}[htb]
    \centering
    \includegraphics[width=0.7\linewidth]{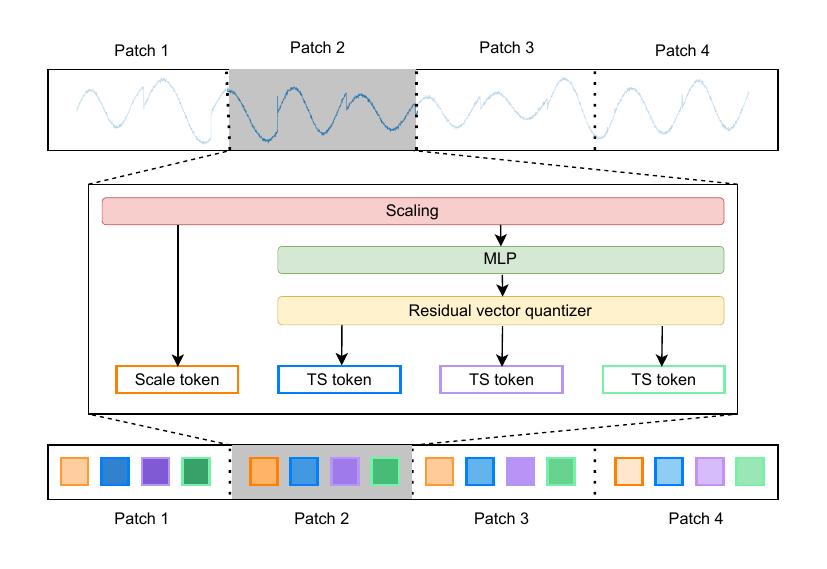}
    \caption{\small The figure shows the tokenization process for time series patches, with Patch 2 highlighted as an example. Each input patch is processed through two parallel pathways: a scaling operation that generates a single scale token capturing magnitude information, and an MLP followed by a residual vector quantizer that produces three time series (TS) tokens encoding temporal patterns. This dual-pathway approach results in four tokens per patch, enabling comprehensive representation of both amplitude and temporal characteristics.}
    \label{fig:diagram:ts-tokenizer}
\end{figure}

The RVQ-VAE is trained on a large and diverse set of general time series data to learn robust representations. 
To ensure training stability and effective codebook utilization, we incorporate several techniques during the pre-training phase, including the rotation trick~\citep{fifty2024restructuring}, dead code expiration~\citep{dhariwal2020jukebox,zeghidour2021soundstream}, k-means initialization~\citep{arthur2006k}, and a commitment loss weighting factor ($\beta$)~\citep{vqvae}.

\paragraph{LLM Integration}
The core of our reasoning framework is a standard decoder-only pre-trained LLM. For our experiments, we default to using Qwen3-8B~\citep{qwen3technicalreport}. The primary modification to the LLM is the extension of its vocabulary to include the new tokens generated by the time series tokenizer, including both time series and scale tokens. This allows the LLM to seamlessly process and generate sequences containing both natural language and time series information.

\subsection{Training Methodology}

The training of COUNTS proceeds in two distinct phases: an initial Supervised Fine-Tuning (SFT) phase to adapt the LLM to time series data and tasks, followed by a Reinforcement Learning (RL) phase to specifically cultivate chain-of-thought reasoning capabilities.

\paragraph{SFT Phase}
The SFT phase is crucial for teaching the LLM to understand and utilize the newly introduced time series tokens, and to perform basic time series analysis tasks. During this phase, the model is trained on a diverse mixture of synthetic and real-world time series analysis tasks that are converted into a prompt-response format. These tasks include interleaved sequences of time-series tokens and corresponding textual descriptions or rationales.
The SFT phase begins with a warm-up period where only the embeddings for the time series tokens are trained, keeping the rest of the LLM parameters frozen. This helps to gently align the new token representations with the LLM's existing knowledge. Following the warm-up, full fine-tuning of the entire model is performed for the remainder of this phase, which spans approximately 10 billion tokens.

\paragraph{Reinforcement Learning Phase}
Building upon the foundation laid by the SFT phase, the RL phase aims to explicitly train the LLM to generate step-by-step reasoning (chain-of-thought) before arriving at a final answer for time series analysis problems. The model initialized from the SFT phase is further trained using the \textbf{D}ecouple Clip and \textbf{D}ynamic s\textbf{A}mpling \textbf{P}olicy \textbf{O}ptimization (DAPO) algorithm~\citep{yu2025dapo}, a variant of the Group Relative Policy Optimization (GRPO) algorithm~\citep{shao2024deepseekmath}. DAPO has shown to improve token efficiency and prevent artificial inflation of response length, particularly for incorrect outputs, by removing length and standard deviation normalization terms from the advantage estimation. In ablation studies (\cref{sec:results:ablation}) we find that it outperforms GRPO in our setting.

The RL phase focuses on tasks with verifiable answers, primarily forecasting, classification, and multiple-choice question answering (MCQ), using real-world data. The LLM generates responses sequentially, and a reward is provided at the end of the generation process. The reward signal is composite, consisting of two components:
\begin{enumerate}
    \item \textbf{Correctness Reward:} This reward is task-dependent. For classification and MCQ tasks we look for an exact match between the prediction and target (allowing for differences in formatting). For forecasting tasks, this reward uses the Symmetric Mean Absolute Percentage Error (SMAPE).
    \item \textbf{Formatting Reward:} To encourage the desired CoT output structure, a formatting reward is used. This reward incentivizes the model to produce its thought process enclosed within \texttt{<think></think>} XML tags and the final answer within \texttt{<answer></answer>} tags, with exactly one think and one answer block expected. 
    Partial credit is given for formats that are partially correct to accelerate the initial learning process.
\end{enumerate} 

During RL training, the model is presented with prompt-answer pairs. The GRPO algorithm then adjusts the policy of the LLM to maximize the expected reward, thereby guiding the model to discover and refine text generation strategies that lead to correct answers and well-structured reasoning steps. This process enables the LLM to learn explicit CoT behavior tailored to time series analysis tasks.

\subsection{RL Implementation Details}
\label{sec:appendix:rl_details}

We train the RL phase using DAPO with AdamW (peak learning rate $1 \times 10^{-5}$, cosine decay)
and gradient norm clipping at 1.0.
For each prompt we sample $G = 32$ rollouts with a mini-batch size of 512 and a maximum sequence
length of 8192 tokens.
The number of RL steps is task-specific: 3000 for ECG-QA, 2000 for CiK forecasting, and 1000 for
UCR classification.
The composite reward sums a task-dependent correctness signal (exact match for classification and
QA; $2 - \text{SMAPE}$ for forecasting) with a formatting reward that incentivizes the expected
\texttt{<think>}\,/\,\texttt{<answer>} structure.
Training uses fully sharded data parallelism across 8 NVIDIA H100 GPUs, implemented with PyTorch and verl.

\section{Results}

We evaluate COUNTS across three diverse time series analysis tasks that require different forms of reasoning: medical signal interpretation through question-answering, contextual forecasting that integrates textual information, and few-shot classification on unseen datasets. These experiments demonstrate that explicit chain-of-thought reasoning, learned through reinforcement learning, significantly enhances performance on complex time series tasks that traditional numerical models struggle with.

\paragraph{Experimental Setup}

We primarily compare against general-purpose LLMs (Gemini 2.5 Pro, o4-mini) because our target tasks require both time series and text inputs or outputs, a combination that few specialized methods can handle. Unless otherwise noted, the LLMs use text-based encodings of the time series; we do not evaluate a text-encoded variant of COUNTS, as training on the resulting long sequences (often exceeding 50,000 tokens) is prohibitively expensive. Where available, we also compare with task-specific models or standard time series methods. All selected baselines were state-of-the-art at the time of our experiments.

To isolate the contribution of our RL training framework from our architecture and pretraining, we compare the full COUNTS model against two ablations: a pre-RL checkpoint that has learned to understand time series but has not been trained on the benchmark datasets, and (for CiK) an SFT-only variant trained on the relevant data without reinforcement learning. In the results tables, the ``\firenospace'' icon indicates that a model was trained on the benchmark dataset, while ``\icenospace'' indicates zero-shot or few-shot evaluation; methods not trained on a benchmark receive a few in-context examples. Although no standardized benchmarks exist for time series reasoning, we have selected high-quality, challenging tasks that reflect potential real-world applications.

\subsection{Question-Answering}

To assess COUNTS's ability to answer challenging questions about time series signals in specialized domains, we use the ECG-QA dataset~\citep{oh_ecg-qa_2023}, a collection of electrocardiogram signals with clinical question-answer pairs. This dataset is particularly difficult because it requires both medical knowledge and detailed time series analysis. Correctly answering questions involves extracting and analyzing temporal features like RR intervals and QRS complexes, then combining them into clinical judgments.
We evaluate on three question types of increasing difficulty: S-Verify questions ask whether a specific attribute is present in the ECG signal; S-Choose questions require selecting which of two attributes best describes the ECG (along with ``both'' or ``neither''); and S-Query questions demand listing all attributes present or identifying specific leads exhibiting particular characteristics.

Table~\ref{tab:qa_results_ecgqa} presents our results against both domain-specific models (M$^3$AE, Q-Heart, ECG-LM) trained explicitly for ECG analysis and general-purpose LLMs. Existing approaches demonstrate significant limitations: specialized models achieve reasonable performance on simpler S-Verify questions but struggle with the more open-ended S-Query tasks, while general-purpose LLMs perform poorly across all question types despite their strong reasoning capabilities in other domains. This performance gap highlights the challenge of combining numerical time series understanding with complex reasoning.

Our approach proceeds in two stages. During the SFT phase, we include a portion of the ECG-QA training data alongside our broader time series corpus, enabling the model to learn basic ECG pattern recognition and medical terminology. At this stage, using few-shot prompting on the test set, COUNTS achieves 60.0\% average accuracy, comparable to the best specialized models but still limited, particularly on S-Query questions (45.2\%).
We then apply reinforcement learning for 3000 steps, using exact match rewards for answer correctness and formatting rewards to encourage proper chain-of-thought structure. This RL training dramatically improves performance to 66.5\% average accuracy, with particularly striking gains on the challenging S-Query questions (53.9\% vs. 45.2\%), surpassing all existing methods by a substantial margin. The 8.7 percentage point improvement on S-Query tasks, where the model must enumerate multiple attributes or identify specific leads, demonstrates that RL-trained chain-of-thought reasoning enables more systematic exploration and verification of complex temporal patterns rather than relying solely on pattern matching.

\begin{table}[h]
\caption{Question-answering exact-match accuracy comparison on ECG-QA by question type. Methods are separated into specialized methods for ECG-QA, general-purpose LLMs, and our methods. The best performance is in \textbf{bold} and the second best is \underline{underlined}.}
\label{tab:qa_results_ecgqa}
\centering
\begin{tabular}{l|rrrr}
    \hline
                             & \multicolumn{4}{c}{\textbf{Question Type}}                    \\
    \textbf{Method} &
      \multicolumn{1}{l}{\textit{S-Verify}} &
      \multicolumn{1}{l}{\textit{S-Choose}} &
      \multicolumn{1}{l}{\textit{S-Query}} &
      \multicolumn{1}{l}{\textit{Average}} \\ \hline
    \fire M$^3$AE                  & 74.6          & 57.1          & 41.0          & 57.6          \\
    \fire Q-Heart                  & \textbf{90.9} & {\ul{60.3}}    & 32.9          & {\ul{61.4}}    \\
    \fire ECG-LM                   & 75.8          & 57.4          & 39.9          & 57.7          \\
    \hline
    \ice Gemini 2.5 Pro           & 55.7          & 30.4          & 17.6          & 30.9          \\
    \ice o4-mini                  & 50.0          & 47.8          & 26.5          & 45.9          \\
    \hline
    \ice ChatTime \ice & 50.1 & 30.9 & 15.5 & 30.5 \\
    \ice ChatTs \ice & 54.4 & 38.2 & 19.1 & 37.2 \\
    \hline
    \ice COUNTS (Pre-RL) & 78.1          & 58.9          & {\ul{45.2}}    & 60.0          \\
    \fire \textbf{COUNTS}            & \textbf{90.9} & \textbf{64.0} & \textbf{53.9} & \textbf{66.5} \\ \hline
    \end{tabular}%
\end{table}

\subsection{Contextual Forecasting}

While traditional forecasting focuses solely on numerical patterns, many real-world prediction tasks require integrating textual context that fundamentally alters the forecasting problem. We evaluate COUNTS's ability to combine numerical time series analysis with contextual reasoning using the Context Is Key (CiK) dataset~\citep{williams2024contextkeybenchmarkforecasting}, which contains 71 realistic forecasting tasks spanning 7 domains specifically designed to require understanding and integrating textual information for successful prediction.

The CiK benchmark reveals a key weakness in traditional forecasting methods: they cannot use contextual information beyond the raw numbers. This context might explain what drives the process, specify constraints, or reveal relationships not visible in the data alone.
Table~\ref{tab:results:cik} demonstrates this challenge starkly -- even powerful traditional models like XGBoost achieve only 76.8\% SMAPE.

Frontier LLMs with strong reasoning capabilities also struggle on this benchmark. Models like o4-mini achieve 72.6\% SMAPE when provided with textual representations of the time series and the relevant context, despite their demonstrated reasoning prowess in other domains. This reveals a complementary failure mode: while these models excel at high-level reasoning, their numerical perception and quantitative prediction capabilities remain insufficient for precise time series forecasting. The challenge lies not just in reasoning about context or analyzing numbers, but in seamlessly integrating both modalities.

Our base COUNTS model, trained only on general time series data, achieves 68.7\% SMAPE, meaning it is already competitive with larger frontier models despite its much smaller size. This improvement stems from COUNTS's superior numerical perception. When we perform further supervised fine-tune specifically on the CiK dataset, performance improves to 61.7\% SMAPE, demonstrating the value of task-specific adaptation. However, the most dramatic gains come from reinforcement learning.

We train COUNTS using RL for 2000 steps with a reward based on forecasting accuracy, specifically using $2 - \text{SMAPE}$ as the correctness reward (where SMAPE is calculated as $\frac{1}{n} \sum_{i=1}^{n} \frac{2 |y_i - \hat{y}_i|}{|y_i| + |\hat{y}_i|}$). This RL training yields a remarkable improvement, achieving a new state-of-the-art SMAPE of 54.5\%, an 18.1 percentage point improvement over the previous best result, surpassing both specialized and frontier models.
The substantial gap between supervised fine-tuning (61.7\%) and RL training (54.5\%) suggests that the explicit reasoning strategies learned through reinforcement learning are fundamentally more effective than the implicit pattern recognition acquired through supervised learning alone.

\begin{table}[h]
\caption{Contextual forecasting performance on the Context Is Key benchmark. For COUNTS, the ``Pre-RL'' variant is our model trained on general time series data but not CiK, and the ``SFT'' model is additionally fine-tuned on the CiK dataset.}
\label{tab:results:cik}
    \centering
    \small
    \begin{tabular}{l|rr}
    \hline
    \textbf{Model}    & \textbf{SMAPE (\%)} $\downarrow$ & \textbf{MASE (\%)} $\downarrow$ \\ \hline
    \fire Linear Regression & 75.4           & 101.5         \\
    \fire XGBoost           & 76.8           & 80.2          \\
    \fire ARIMA             & 90.7           & 134.4         \\
    \fire ETS               & 110.0          & 204.2         \\
    \hline
    \ice Gemini 2.5 Pro    & 90.8           & 98.1          \\
	\ice o4-mini           & 72.6           & 70.5          \\
    \hline
    \ice ChatTime & 70.1 & 71.4 \\
    \ice ChatTs & 68.4 & 69.2 \\
    \hline
    \ice COUNTS (Pre-RL)       & 68.7           & 70.1          \\
    \fire COUNTS (SFT)    & {\ul{61.7}}    & {\ul{64.8}}          \\
    \fire COUNTS              & \textbf{54.5}  & \textbf{58.1} \\ \hline
    \end{tabular}
\end{table}

\subsection{Classification}

Beyond domain-specific tasks, we evaluate COUNTS's ability to develop generalizable reasoning strategies using the UCR Time Series Classification benchmark~\citep{UCRArchive2018}, a standard collection of 128 datasets across diverse domains. While individual UCR datasets suit traditional methods well, we investigate whether COUNTS can learn few-shot classification on new datasets through reasoning rather than memorizing dataset-specific patterns.

Our experimental design specifically tests generalization capability. We completely exclude the UCR datasets from supervised fine-tuning, ensuring the model has no prior exposure to these classification tasks. For reinforcement learning, we randomly hold out 32 datasets as a test set and train on the remaining 96 datasets mixed together for 1000 steps. This setup prevents the model from learning dataset-specific strategies and instead encourages it to develop general reasoning approaches for few-shot time series classification.

During evaluation, we use few-shot prompting where the model receives 4-10 labeled examples per class before predicting on test instances. This approach is essential because without examples, the model would need prior knowledge of each dataset's patterns and class definitions. Instead, few-shot prompting lets the model examine examples, identify key patterns, and apply this understanding to new cases -- similar to how human experts approach unfamiliar classification tasks.

Table~\ref{tab:results:classification} presents results on the 32 held-out datasets. Traditional models trained individually on each dataset achieve strong performance, with XGBoost reaching 68.6\% mean accuracy, which is unsurprising given these methods can fully optimize for each specific task. General-purpose LLMs struggle significantly, achieving only 40.9\% and 44.8\% accuracy respectively, highlighting their limitations in few-shot time series analysis.

Our pre-RL model achieves 53.5\% mean accuracy, demonstrating reasonable few-shot learning capability from the general time series training. However, RL training produces a substantial improvement to 60.1\% mean accuracy, a 6.6 percentage point gain that approaches the performance of Random Forest (61.0\%) despite never seeing these datasets during training.

While COUNTS does not surpass dataset-specific learning, the strong improvement from RL training validates our core hypothesis: explicit chain-of-thought reasoning enhances a model's ability to tackle novel time series problems by learning generalizable analytical strategies rather than memorizing task-specific patterns. This capability is invaluable for real-world applications where collecting large labeled datasets for every new classification task is impractical.

\begin{table}[h]
\caption{Classification accuracy averaged over a selection of 32 held-out datasets in the UCR Time Series Classification benchmark. Models in italics were not trained on these datasets.}
\label{tab:results:classification}
    \centering
    \small
    \begin{tabular}{l|rr}
    \hline
    \textbf{Metric}           & \textbf{Mean Acc.} $\uparrow$ & \textbf{Median Acc.}  $\uparrow$\\ \hline
    \fire Logistic Regression       & 0.452              & 0.481                \\
    \fire Random Forest            & {\ul{0.610}}        & {\ul{0.639}}          \\
    \fire XGBoost                   & \textbf{0.686}     & \textbf{0.662}       \\
    \hline %
    \ice Gemini 2.5 Pro   & 0.409              & 0.427                \\
    \ice o4-mini          & 0.448              & 0.457                \\
    \hline
    \ice ChatTs             & 0.411              & 0.413                \\
    \ice ChatTime           & 0.454              & 0.442                \\
    \hline
    \ice COUNTS (Pre-RL)  & {\ul{0.535}}     & {\ul{0.537}}        \\
    \ice COUNTS             & \textbf{0.601}    & \textbf{0.589}       \\ \hline
    \end{tabular}%
\end{table}

\subsection{Ablation Studies} \label{sec:results:ablation}

Table~\ref{tab:ablation} analyzes component contributions of COUNTS on ECG-QA. Key findings are (1) model scale matters -- upgrading from Qwen3-4B (used for the baselines) to Qwen3-8B (used in the main experiments) parameters yields +5.1pp. (2) RL algorithm choice is critical -- DAPO outperforms plain GRPO by 9.0pp and GSPO by 5.9pp. (3) An RVQ-VAE based tokenizer outperforms a VQ-VAE tokenizer by 7.4pp, indicating that the improved fidelity and split codebook helps. (4) Larger patches (64 vs 16 points) improve performance by 5.7pp, suggesting that a higher information density in the encodings can help.

\begin{table}[htb]
\centering
\caption{Ablation study for COUNTS using mean accuracy on ECG-QA.}
\label{tab:ablation}
\begin{tabular}{l|l|r}
\hline
\textbf{Ablation Category}   & \textbf{Value}      & \textbf{Accuracy} $\uparrow$  \\ \hline
Baseline   &            & 61.4                     \\
LLM & Qwen3 8B & 66.5 \\
RL method  & GRPO & 57.5 \\
RL method  & GSPO       & 60.6                     \\
TS encoder & VQ-VAE     & 59.1                     \\
Patch size & 16         & 60.8 \\
Group size & 16         & 60.4 \\
\hline
\end{tabular}%
\end{table}

\section{Discussion}

Our RL approach successfully teaches LLMs explicit chain-of-thought reasoning for time series.
Across all benchmarks, RL yielded consistent improvements (6.6–18.1 points), underscoring its value as a reasoning amplifier.
These consistent gains reveal a fundamental reasoning gap that supervised learning alone cannot bridge, establishing time series as a third viable domain for RL-based reasoning alongside mathematics and code generation.
The strong correlation between task complexity and RL benefits is particularly telling. Our largest improvements occur precisely where sophisticated reasoning matters most: open-ended ECG analysis, contextual forecasting with constraints, and few-shot classification. This pattern exposes a key limitation of current approaches -- while traditional models excel at statistical pattern recognition, they struggle when solutions require reasoning about relationships, integrating domain knowledge, or adapting to novel contexts.
These findings suggest a useful framework distinguishing procedural knowledge (how to analyze patterns) from declarative knowledge (domain-specific facts). Our results indicate RL excels at teaching the former while supervised learning handles the latter, pointing toward hybrid training approaches that explicitly leverage both learning modes.

\paragraph{Limitations}

Our approach faces three main limitations. First, RL training remains task-specific, with separate reward metrics for ECG, forecasting, and classification. Second, our tokenizer currently handles univariate inputs, requiring channel splitting for multivariate signals; joint quantization or scale-fusion would better capture cross-variable structure. Third, the computational cost for training is substantial compared to traditional time series models. Despite these, the consistent gains across tasks highlight that RL-based reasoning offers a robust foundation for future unified models.

\clearpage

\bibliographystyle{unsrtnat}
\bibliography{references_manual,references}

\clearpage
\appendix

\section{Benchmark Dataset Details}

\subsection{ECG-QA}

To test COUNTS's reasoning about complex temporal patterns in specialized domains, we use the ECG-QA dataset~\citep{oh_ecg-qa_2023}, a collection of electrocardiogram signals with clinical question-answer pairs. This dataset is particularly challenging because it requires both medical knowledge and detailed time series analysis. Questions involve extracting temporal features like RR intervals and QRS complexes, then combining them into clinical judgments. For example, answering "has the PR interval normalized compared to the previous ECG?" requires identifying waveform components, measuring durations, and making comparisons -- combining perception with logical reasoning.

The dataset encompasses six clinically grounded attribute families, specifically, SCP codes, noise types, myocardial infarction stages, extra systoles, heart axis measurements, and numeric intervals derived from raw time series, with many questions being lead-specific. We evaluate on three question types of increasing difficulty: S-Verify questions ask whether a specific attribute is present in the ECG signal; S-Choose questions require selecting which of two attributes best describes the ECG (with ``both'' or ``neither'' as valid options); and S-Query questions demand listing all attributes present or identifying specific leads exhibiting particular characteristics. This progression from binary verification to open-ended enumeration tests increasingly sophisticated reasoning capabilities.

\section{Data Samples}

\subsection{Example 1}

\textbf{Input:}
Answer the following question about the given ECG signal. The given signal is a 10 second clip of a 12-lead ECG signal recorded at 250Hz. Respond with one of the possible answers and nothing else.

\textbf{Question:} Which diagnostic symptom does this ECG show, subendocardial injury in inferolateral leads or long qt-interval, including uncertain symptoms?

\textbf{Possible Answers}
\begin{enumerate}
    \item complete left bundle branch block
    \item complete right bundle branch block
    \item digitalis effect
    \item first degree av block
    \item incomplete left bundle branch block
    \item incomplete right bundle branch block
    \item ischemic in anterior leads
    \item ischemic in anterolateral leads
    \item ischemic in anteroseptal leads
    \item ischemic in inferior leads
    \item ischemic in inferolateral leads
    \item ischemic in lateral leads
    \item left anterior fascicular block
    \item left atrial overload/enlargement
    \item left posterior fascicular block
    \item left ventricular hypertrophy
    \item long qt-interval
    \item myocardial infarction in anterior leads
    \item myocardial infarction in anterolateral leads
    \item myocardial infarction in anteroseptal leads
    \item myocardial infarction in inferior leads
    \item myocardial infarction in inferolateral leads
    \item myocardial infarction in inferoposterior leads
    \item myocardial infarction in inferoposterolateral leads
    \item myocardial infarction in lateral leads
    \item myocardial infarction in posterior leads
    \item non-diagnostic t abnormalities
    \item non-specific intraventricular conduction disturbance (block)
    \item non-specific ischemic
    \item non-specific st changes
    \item none
    \item right atrial overload/enlargement
    \item subendocardial injury in anterolateral leads
    \item subendocardial injury in anteroseptal leads
    \item subendocardial injury in inferior leads
    \item subendocardial injury in inferolateral leads
    \item subendocardial injury in lateral leads
\end{enumerate}

\textbf{ECG signals}
\begin{itemize}
    \item lead I: [ECG Image 1]
    \item lead II: [ECG Image 2]
    \item lead III: [ECG Image 3]
    \item lead aVR: [ECG Image 4]
    \item lead aVL: [ECG Image 5]
    \item lead aVF: [ECG Image 6]
    \item lead V1: [ECG Image 7]
    \item lead V2: [ECG Image 8]
    \item lead V3: [ECG Image 9]
    \item lead V4: [ECG Image 10]
    \item lead V5: [ECG Image 11]
    \item lead V6: [ECG Image 12]
\end{itemize}

\texttt{<|thought>}\textbf{Analysis of a 12-Lead ECG: Expert Opinion}
Right, let's dive into this ECG. My initial impression is that the heart rate is quite elevated, and I'm seeing some ST segment changes that warrant a closer look.
\begin{enumerate}
    \item \textbf{General ECG Assessment}: I'm starting by establishing the basics. The rhythm seems regular, but the high heart rate, estimated around 214 bpm, immediately catches my attention. I'm noting a normal axis, which is always a good starting point. The PR interval appears within normal limits, though a bit challenging to precisely measure due to some noise. The QRS duration is also within the normal range, so no bundle branch block is evident.
    \item \textbf{QT Interval Evaluation}: Now, the crucial part: the QT interval. This is where things get interesting, and a little complex, due to the rapid heart rate.  I've attempted a few measurements, and the initial calculated QT intervals seemed short. But, because the heart rate is so high, a straight measurement is unreliable. I've had to use Bazett's formula to correct for heart rate, and even then, at these speeds, Bazett can be inaccurate. I also tried Fridericia's formula, which is considered more accurate at faster heart rates, and that gave me a more normal corrected QT. I then found a few other formulas, including Hodges and Framingham.  This is giving me some conflicting results. Hodges gives a very high QTc, but the consensus on the other formulas and the visible T wave is that the QT is not definitively long, especially given the rate. So, while I'm not ruling out a slight borderline prolongation, I'm leaning against "long QT-interval" as a primary diagnosis based on these readings.
    \item \textbf{Subendocardial Injury Investigation}: Now, let's turn to the other potential issue: "subendocardial injury in inferolateral leads." This is where the ST segments become crucial. I'm focusing on the inferior leads (II, III, aVF) and the lateral leads (I, aVL, V5, V6), where I'm looking for ST depression. Bingo! I'm seeing clear, downsloping ST depression in leads II, III, aVF, V5, and V6. That's a strong indicator. I am also seeing ST elevation in aVR, which indicates a reciprocal change, that goes along with diffuse subendocardial ischemia.
    \item \textbf{Comparing the Two Options:} So, where does this leave us? The "long QT-interval" diagnosis is questionable, given the variable results of different correction methods at very high heart rates. However, the evidence for "subendocardial injury in inferolateral leads" is compelling and unambiguous. I'm seeing clear ST depression in multiple leads, indicating myocardial ischemia. Given that the option I can select from involves a diagnosis, I must choose the best-supported.
    \item \textbf{Final Verdict:} The ECG is showing a pattern of widespread subendocardial ischemia, with clear ST depression in inferior and lateral leads. The presence of ST elevation in aVR adds to the evidence of a broader injury pattern. While a borderline prolonged QT interval is \textbf{possible}, it is not the primary finding I'm seeing here. So, the most fitting answer is the ``subendocardial injury in inferolateral leads''.
    \item \textbf{Additional Considerations:} The very high heart rate and the widespread nature of the ST depression also suggest a potential for ischemia, possibly a type 2 MI, and the need for immediate clinical attention. 
\end{enumerate}
\texttt{<|thought>}

\textbf{Detailed Analysis}
The user wants me to analyze the provided 12-lead ECG and determine which of the two specified conditions is present: "subendocardial injury in inferolateral leads" or "long qt-interval".

\begin{enumerate}
    \item \textbf{Rhythm and Rate}: The ECG shows a regular, narrow-complex tachycardia. Heart rate calculation:
\begin{align}
\text{R-R interval} &= \frac{70 \text{ samples}}{250 \text{ Hz}} = 0.28 \text{ seconds} \\
\text{Heart Rate} &= \frac{60}{0.28} \approx 214 \text{ bpm}
\end{align}

This represents supraventricular tachycardia (SVT).

    \item \textbf{Axis}: Lead I appears roughly isoelectric/slightly positive. Lead aVF is positive. The axis is likely normal, around $+90$ degrees.
    \item \textbf{ST-T Wave Morphology}
\begin{itemize}
    \item \textbf{Inferior Leads (II, III, aVF):} All show significant, downsloping ST-segment depression
    \item \textbf{Lateral Leads (I, aVL, V5, V6):} V5 and V6 show clear ST-segment depression
    \item \textbf{Anterior/Septal Leads (V1-V4):} Also show significant ST depression
    \item \textbf{Lead aVR:} Shows prominent ST-segment elevation
\end{itemize}

    \item \textbf{QT Interval Analysis}
    At heart rate $> 210$ bpm, accurate QT measurement and correction is challenging. Measurement in Lead V3:
    \begin{align}
    \text{QT interval} &= \frac{55 \text{ samples}}{250 \text{ Hz}} = 0.22 \text{ seconds} \\
    \text{QTc (Bazett)} &= \frac{0.22}{\sqrt{0.28}} \approx 0.416 \text{ s (416 ms)} \text{ - borderline} \\
    \text{QTc (Fridericia)} &= \frac{0.22}{\sqrt[3]{0.28}} \approx 0.336 \text{ s (336 ms)} \text{ - normal}
    \end{align}
    
    Given the ambiguity and normal results from reliable formulas, ``long QT-interval'' diagnosis is uncertain.
\end{enumerate}

\textbf{Final Decision}
\begin{itemize}
    \item The evidence for subendocardial injury, indicated by significant ST depression in the inferolateral leads (and other leads), is very strong and visually obvious.
    \item The evidence for a long QT interval is weak and highly dependent on the correction formula used at this very fast heart rate.
    \item Therefore, the most accurate and certain diagnosis based on the provided ECG is subendocardial injury.
\end{itemize}

The ECG shows clear ST depression in leads II, III, aVF, V5, and V6, which constitute the inferolateral leads.

\textbf{Final Decision}
The evidence for subendocardial injury (significant ST depression in inferolateral leads) is strong and visually obvious. The evidence for long QT interval is weak and highly dependent on correction formula used at this fast heart rate.

The ECG shows clear ST depression in leads II, III, aVF, V5, and V6, which constitute the inferolateral leads.

\boxed{\text{(36) subendocardial injury in inferolateral leads}}

\section{Training Data}

Table \ref{tab:datasets} summarizes the key statistics for all datasets used in the supervised fine-tuning stage of COUNTS. Our training corpus encompasses nearly 27 billion total tokens across 20 distinct datasets, spanning diverse domains including medical signals, financial data, environmental measurements, and industrial sensors. The datasets support multiple task types including forecasting, classification, anomaly detection, segmentation, and question-answering, enabling COUNTS to develop comprehensive time series analysis capabilities.

\begin{table*}[h]
\centering
\tiny
\begin{tabular}{lllrrrr}
\toprule
Name & Domain & Tasks & Total Tokens & Response Tokens & Text/TS Tokens & Samples \\
\midrule
GIFT-Pretrain & Multi-Domain & Forecasting & 7,951,924,108 & 3,861,122,236 & (1,990,801,872/5,961,122,236) & 74,805,850 \\
ECG-QA & ECG & Q-A& 1,071,939,892 & 10,119,218 & (468,584,812/603,355,080) & 942,950 \\
News-Stocks & Finance & Forecasting & 124,179,199 & 3,464,920 & (118,277,891/2,436,388) & 61,922 \\
PTBXL-Classification & ECG & Classification & 20,615,930 & 296,710 & (3,577,588/16,741,632) & 21,799 \\
PTBXL-Forecasting & ECG & Forecasting & 22,482,365 & 5,740,733 & (0/16,741,632) & 21,799 \\
PTBXL-Report Gen & ECG & Report Gen. & 23,155,582 & 3,689,075 & (2,724,875/16,741,632) & 21,799 \\
Monster & Multi-Domain & Classification & 1,665,431,147 & 12,045,400 & (929,008,270/736,422,877) & 1,998,336 \\
TTC-Medical & Medical & Mixed & 628,367 & 232,387 & (390,936/5,044) & 330 \\
TTC-Climate & Climate & Mixed & 2,132,627 & 559,584 & (1,563,351/9,692) & 1,320 \\
UCR-TSC & Time Series & Classification & 68,966,571 & 797,925 & (50,634,942/17,533,704) & 60,555 \\
UTSD & Multi-Domain & Forecasting & 78,164,051 & 51,392,965 & (0/26,771,086) & 873,621 \\
Weather5K & Weather & Forecasting & 673,077,250 & 515,506,637 & (665,491,994/7,585,256) & 674,245 \\
UEA & Multi-Domain & Classification & 29,306,232 & 1,821,291 & (16,888,337/10,596,604) & 22,606\\
Child Sleep Detct. & Sensor & Segmentation & 916,281 & 52,061 &(531,027/333,193) & 5,331\\
PSM & Machine & Anomaly & 704,856 & 240,546 & (285,299/179,011) & 220,322 \\
MSL & Spacecraft & Anomaly & 893,631 & 304,969 & (361,708/226,954) & 132,046 \\
SMAP & Spacecraft & Anomaly  & 1,731,267 & 590,829 & (700,751/439,687) & 562,800\\
SWaT & Infrastructure & Anomaly & 6,106,917 & 2,084,106 & (2,471,848/1,550,963) & 944,920\\
SMD & Machine & Anomaly & 6,624,762 & 2,260,831 & (2,681,452/1,682,479) & 1,416,825 \\
Synthetic & Multi-Domain & Mixed & 15,186,959,206 & 2,815,439,507 & (11,267,019,855/3,919,939,351) & 11,273,306 \\
\midrule
\textbf{Total} & \textbf{All} & \textbf{All} & \textbf{26,935,940,241} & \textbf{7,287,761,930} & \textbf{(15,521,996,808/11,340,414,501)} & \textbf{94,062,682} \\
\bottomrule
\end{tabular}
\caption{Comprehensive statistics for all datasets used in TsLLM training, showing the distribution of tokens across text and time series modalities, along with task diversity across domains.}
\label{tab:datasets}
\end{table*}

\end{document}